\documentclass[times, review, 10pt]{elsarticle}

\usepackage{lineno,hyperref}
\usepackage{appendix}
\usepackage{subfigure}
\usepackage{graphicx}
\usepackage{wrapfig}
\usepackage{amsmath,amssymb,amsfonts}
\usepackage{algorithm}
\usepackage{algpseudocode}
\usepackage{multirow}
\usepackage{threeparttable}
\usepackage{bbding}
\usepackage{ulem}
\usepackage{url}
\usepackage{booktabs}
\usepackage{makecell}
\usepackage{bm}
\usepackage{color}
\usepackage{twemojis}
\usepackage{enumitem}
\usepackage{natbib}
\journal{Nuclear Physics B}

\begin{document}

\begin{frontmatter}



\title{A comprehensive survey of oracle character recognition: challenges, benchmarks, and beyond}


\author[int]{Jing Li}
\author[xjtul]{Xueke Chi}
\author[xjtul]{Qiufeng Wang}
\author[Duke]{Kaizhu Huang}
\author[xut]{Da-Han Wang}
\author[aynu]{Yongge Liu}
\author[iacas]{Cheng-Lin Liu}

\address[int]{College of Science \& Technology Ningbo University, Ningbo, China}

\address[xjtul]{School of Advanced Technology, Xi'an Jiaotong-Liverpool University, Suzhou, China}

\address[Duke]{Duke Kunshan University, Suzhou, China}

\address[xut]{Fujian Key Laboratory of Pattern Recognition and Image Understanding, School of Computer and Information Engineering, Xiamen University of Technology, Xiamen, China}

\address[aynu]{Anyang Normal University, Anyang, China}

\address[iacas]{State Key Laboratory of Multimodal Artificial Intelligence Systems, Institute of Automation, Chinese Academy of Sciences, Beijing, China}

\begin{abstract}
Oracle character recognition—an analysis of ancient Chinese inscriptions found on oracle bones—has become a pivotal field intersecting archaeology, paleography, and historical cultural studies. Traditional methods of oracle character recognition have relied heavily on manual interpretation by experts, which is not only labor-intensive but also limits broader accessibility to the general public. 
With recent breakthroughs in pattern recognition and deep learning, there is a growing movement towards the automation of oracle character recognition (OrCR), showing considerable promise in tackling the challenges inherent to these ancient scripts.
However, a comprehensive understanding of OrCR still remains elusive. Therefore, this paper presents a systematic and structured survey of the current landscape of OrCR research. We commence by identifying and analyzing the key challenges of OrCR. Then, we provide an overview of the primary benchmark datasets and digital resources available for OrCR. A review of contemporary research methodologies follows, in which their respective efficacies, limitations, and applicability to the complex nature of oracle characters are critically highlighted and examined.
Additionally, our review extends to ancillary tasks associated with OrCR across diverse disciplines, providing a broad-spectrum analysis of its applications.
We conclude with a forward-looking perspective, proposing potential avenues for future investigations that could yield significant advancements in the field.
\end{abstract}


\end{frontmatter}


\section{Introduction}\label{intro}
\begin{wrapfigure}[12]{r}{0.63\textwidth}
\centering
\vspace{-0.4cm}
\subfigure[]{\rotatebox{-90}{\includegraphics[scale=0.18]{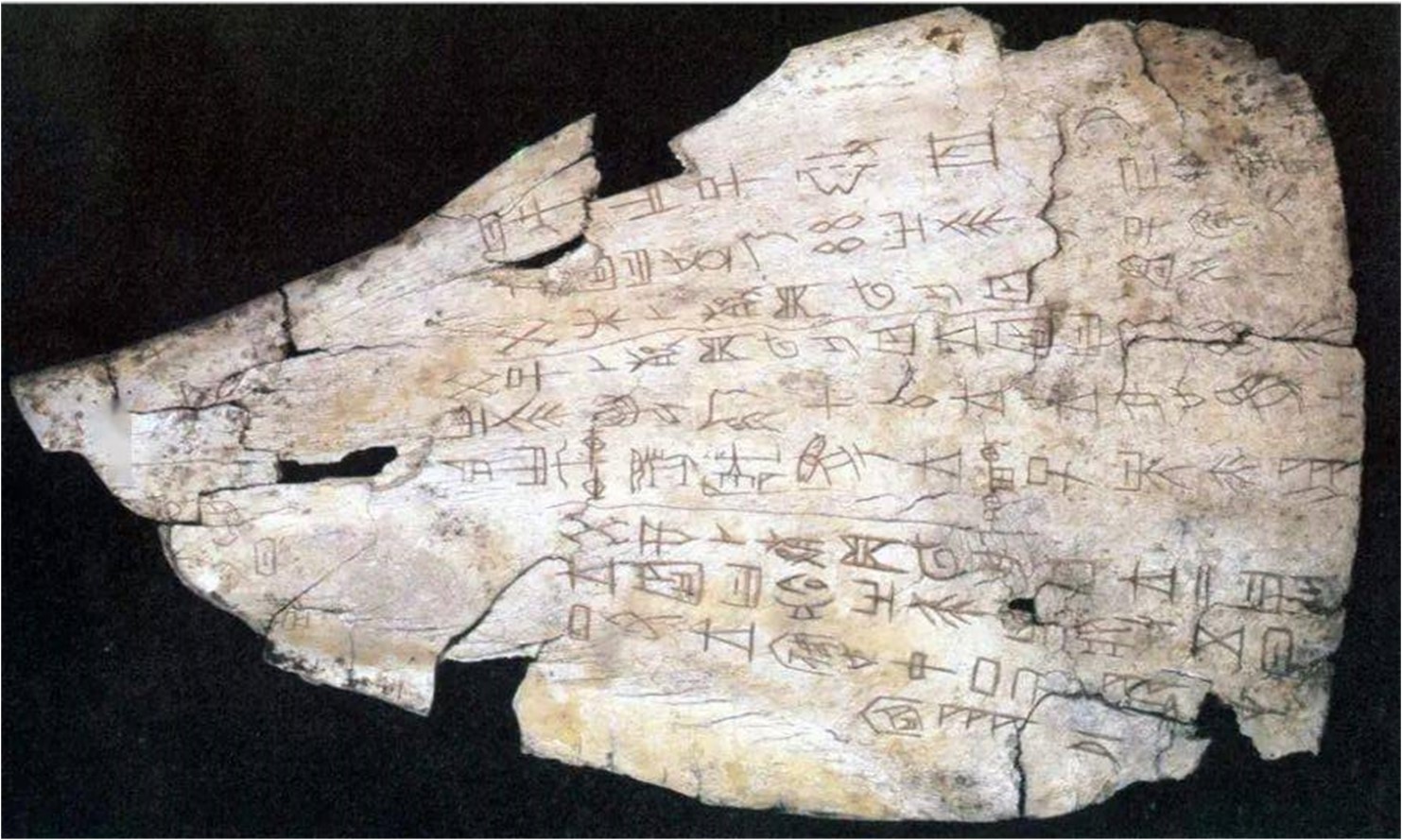}}
\label{fig:intro_bone}}
\quad
\subfigure[]{\rotatebox{-90}{\includegraphics[scale=0.18]{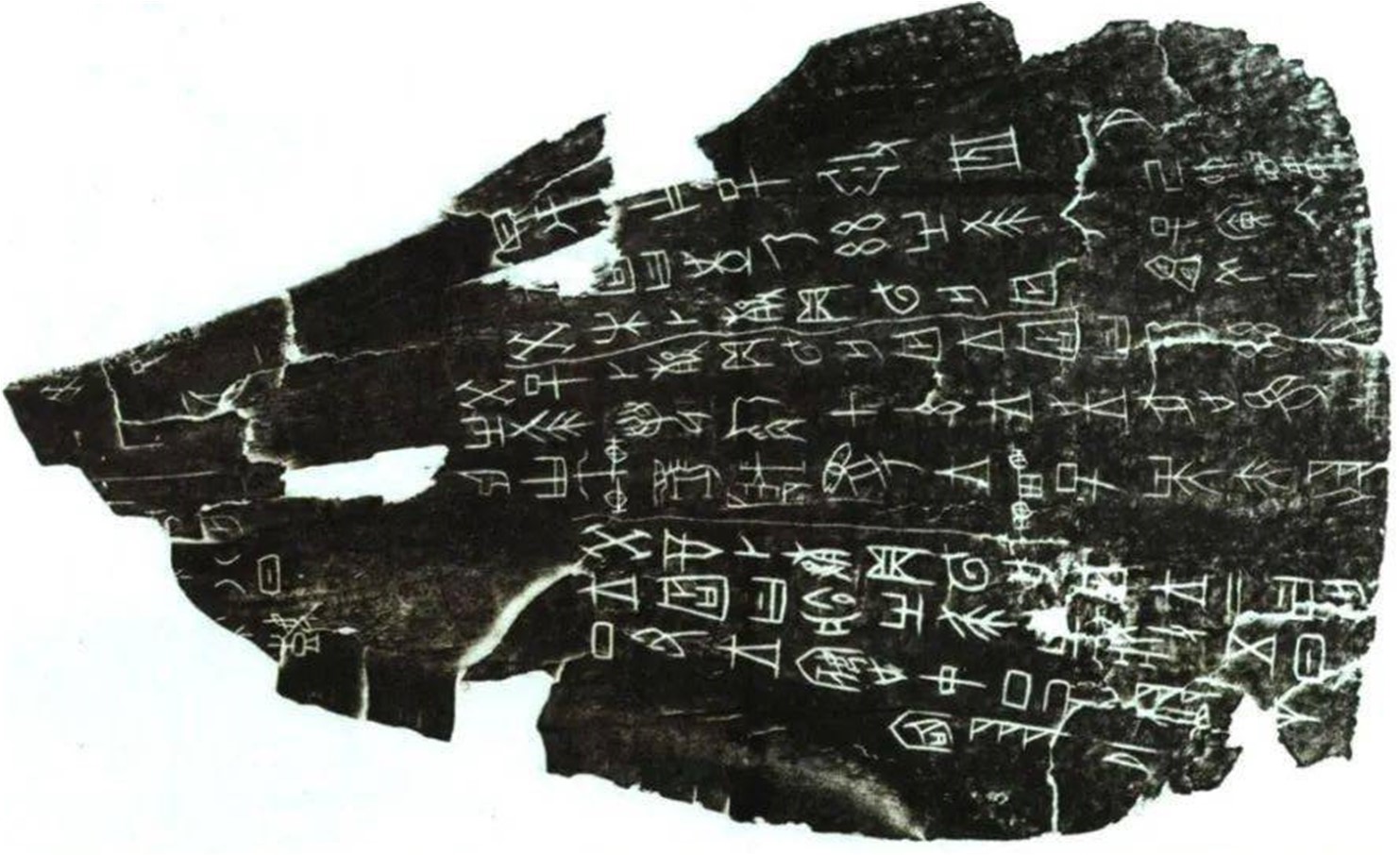}}
\label{fig:intro_rubbing}}
\vspace{-0.3cm}
\caption{Examples of an oracle bone (a) and its corresponding scanned rubbing (b).}
\label{fig:intro}
\end{wrapfigure}

The oracle bone script, dating back about 3,500 years, represents the earliest known mature writing system in China. These characters were typically engraved on turtle shells or animal bones for divination by Shang dynasty rulers. Fig.~\ref{fig:intro_bone} shows an inscribed oracle bone.  Oracle characters carry the history and culture of ancient China, making their study crucial for understanding Chinese writing origins and reconstructing Chinese history. However, interpreting these characters usually requires multidisciplinary expertise in paleography, history, and archaeology, limiting the field to a small number of oracleologists.

Since the first discovery of oracle bones in the late nineteenth century, around 4,500 character classes have been discovered, but merely about 2,200 have been mapped to modern equivalents \cite{OBC306}. Due to the difficulty of manual interpretation, researchers in computer science and oracleology have increasingly explored pattern recognition and deep learning techniques to automate oracle character recognition. These techniques could serve as auxiliary tools for oracleologists while also helping the public understand and read oracle characters.


In computer research, oracle bones are usually converted into rubbings and then scanned into digital images, as illustrated in Fig.~\ref{fig:intro_rubbing}. These images undergo automatic recognition to identify oracle characters. The process is generally divided into two subtasks: oracle character localization and isolated oracle character recognition. 
In this paper, we focus primarily on automating isolated oracle character recognition (OrCR) unless otherwise stated. Based on our analysis, there are three main challenges in OrCR, including writing variability, data scarcity, and low image quality. 

\begin{wrapfigure}[13]{l}{0.55\textwidth}
\centering
\vspace{-0.5cm}
\subfigure[]{\includegraphics[scale=0.35]{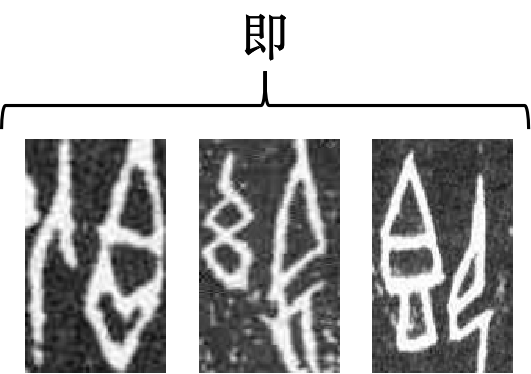}}
\subfigure[]{\includegraphics[scale=0.35]{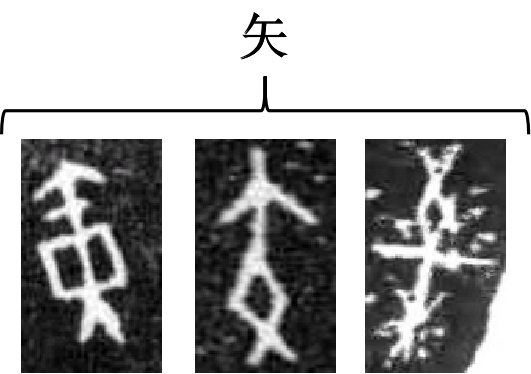}
}
\subfigure[]{\includegraphics[scale=0.35]{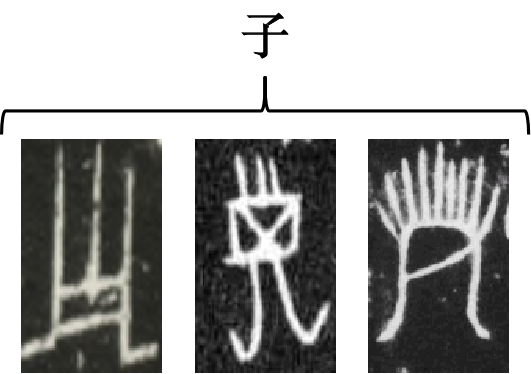}\label{fig:intra-class2}}
\subfigure[]{\includegraphics[scale=0.35]{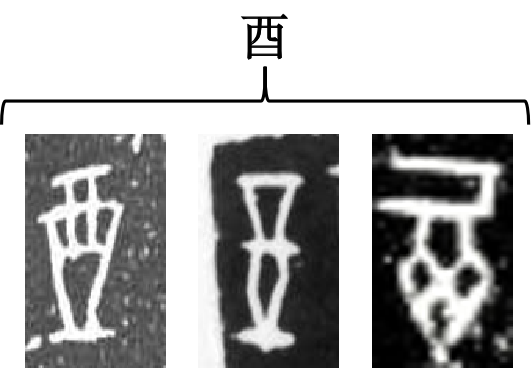}}
\vspace{-0.3cm}
\caption{Illustration of writing variability. Samples different in writing are represented by the same class in modern  Chinese characters (Top).}
\label{fig:intra-class}
\end{wrapfigure}

\begin{figure}[htbp]
\centering
\subfigure[]{\label{fig:12c}
\includegraphics[height=3.8cm]{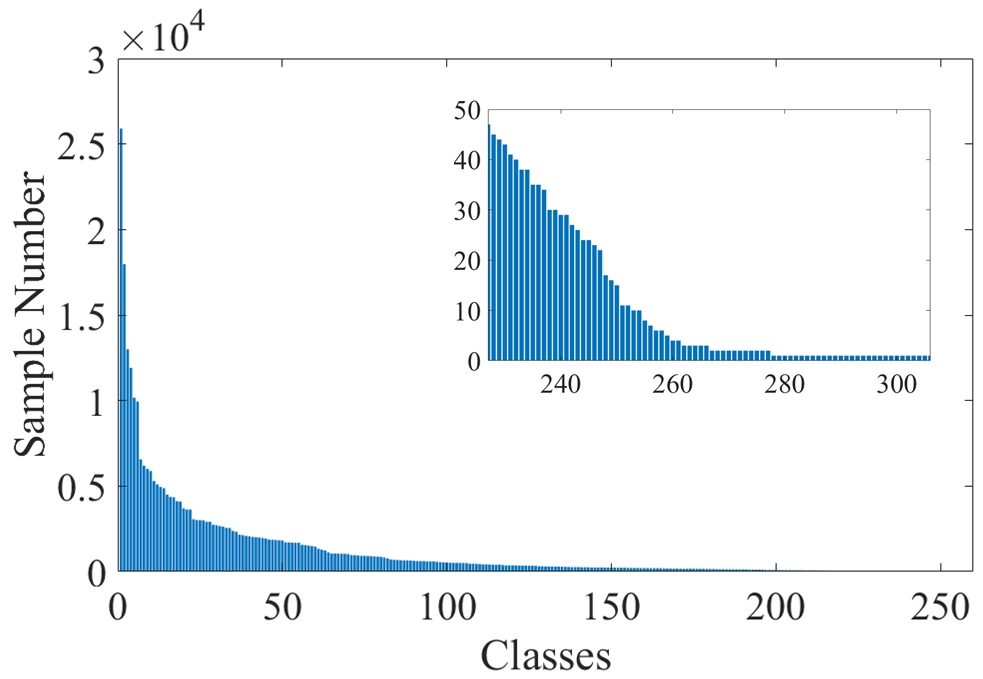}}
\quad
\subfigure[]{\label{fig:12d}
\includegraphics[height=3.8cm]{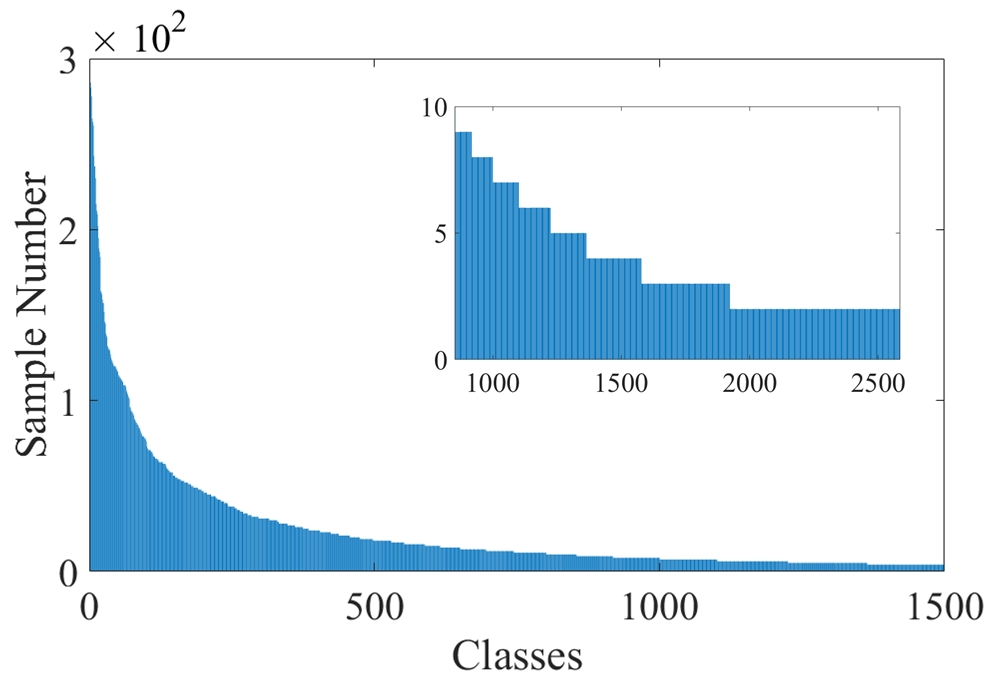}}
\vspace{-0.3cm}
\caption{Class distributions in two oracle datasets. a) OBC306 \cite{OBC306}. b) Oracle-AYNU \cite{icdar2019}.}
\label{fig:data_distribution}
\end{figure}
\begin{wrapfigure}[9]{r}{0.62\textwidth}
\vspace{-0.45cm}
\centering
\subfigure[]{\includegraphics[scale=0.43]{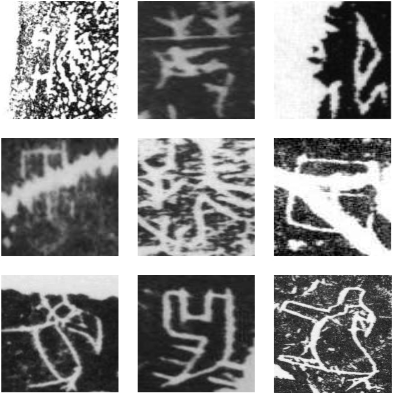}\label{fig:obc}}
\hfill
\subfigure[]{\includegraphics[scale=0.43]{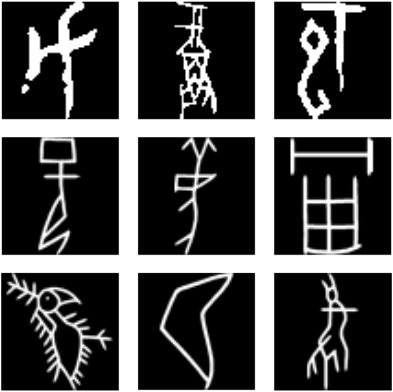}\label{fig:handcopied}}
\vspace{-0.3cm}
\caption{Examples of oracle data. Isolated scanned characters (a) and isolated handprinted characters (b).}
\label{fig:oracle_charcter_examples}
\end{wrapfigure}

First, the development of oracle characters lacked standardized norms, leading to significant variation in character writing. Characters with the same meaning were often written differently across regions and periods, creating multiple variants within a single character class, as shown in Fig.~\ref{fig:intra-class}.  
Second, unlike modern Chinese characters, oracle character samples are rare, causing data scarcity and class imbalance in most benchmark datasets, as illustrated in Fig.~\ref{fig:data_distribution}.
Third, as shown in Fig.~\ref{fig:obc}, oracle bone images often suffer from low quality due to long-term burial and improper excavation. This results in partially missing edges, dense white regions, broken strokes, and cracked bones \cite{OBC306}. Although experts can copy oracle characters by hand to obtain high-quality images, as shown in Fig.~\ref{fig:handcopied}, this process is time-consuming and costly. 
In summary, such three challenges make OrCR far from being fully solved. 

To address the above challenges, significant progress has been made in OrCR, from traditional pattern recognition \cite{zhou1995jia,Meng2017Line} to mainstream deep learning approaches \cite{icdar2019,li2021mix,wang2022unsupervised}. Despite these advancements, to date, there remains a scarcity of literature systematically reviewing OrCR works. 
To fill this gap, we provide an in-depth analysis and comprehensive survey of recent OrCR. To the best of our knowledge, this is the first systematic and structured survey in this field, offering researchers and the community a deeper understanding of the progress in OrCR, mainly on five essential aspects: challenges, datasets and resources, methodology, related tasks, and future works.

In this paper, we first analyze the three key challenges. 
Second, we introduce the most representative oracle datasets and resources, detailing their respective characteristics such as data sources, types, and volumes. To provide a clear understanding of current progress, we also present state-of-the-art performance on each dataset. 
Third, we provide an overview of OrCR methodologies, discussing how each addresses the key challenges.
Fourth, we explore related applications, highlighting OrCR's broader significance. 
Finally, we discuss emerging directions to guide future research.
This paper serves as both an introduction for newcomers and a detailed review for experienced researchers in OrCR.




\section{Challenges}\label{challenges}
Oracle character recognition (OrCR) aims to identify isolated oracle characters from a predefined set of classes. There are three main challenges to be solved, including the intrinsic variability in writing styles, the quantity limitations of available data, and the degraded quality of scanned images.

\subsection{Writing variability}\label{challenges_intra}
In contrast to modern Chinese character forms, the writing forms of oracle characters exhibit significant variations, resulting in large intra-class variance \cite{li2023towards,hu2024component}. This variability stems from the following factors.
\begin{itemize}[topsep=0pt,itemsep=0pt]
    \item \textbf{Influence of Writing Tools and Materials.} Oracle characters were directly carved onto hard turtle shells or animal bones using knives, leading to distinct writing conditions. Factors, such as the sharpness of the knife, the skill of the carver, and the hardness of the bone or shell, contributed to variations in character shapes and strokes, even within the same character class. For instance, some strokes were simplified or omitted due to the difficulty of carving or damage to the materials, producing variant forms.
    \item \vspace{-0.15cm}\textbf{Individual Differences Among Scribes.} Oracle characters were inscribed by diviners or other specialized individuals. Since there was no standardized writing system at that time, scribes might depict the same character in different ways, based on their personal habits or interpretations. For example, a character representing an animal could appear in different variants depending on the scribe's focus. This variability results in multiple forms of the same character.
    \item \vspace{-0.15cm}\textbf{Regional and Temporal Differences.} Oracle characters were used across a wide geographic range and over an extended time period, resulting in substantial writing variations.

\end{itemize}

In summary, oracle characters exhibit a rich diversity of forms. 
According to statistical data from \cite{restudyyinshang}, there are 1,032 groups of variant oracle characters, comprising 3,085 glyphs. For example, Fig.~\ref{fig:intra-class2} displays three variants of one oracle character class ( ``Zi", meaning ``Child"). The first variant has a simple structure with straight lines, while the second is more complex with an eye-like upper part and two curved lines resembling legs, giving it a human-like appearance. The third variant is more pictographic, with several vertical lines in the upper part resembling hair and the lower part forming an arc. Such writing variability poses large intra-class variance, resulting in significant challenges for automatic recognition.

\subsection{Data Scarcity}\label{challenges_im}
The scarcity of oracle character samples poses significant challenges for statistical models, which, in general, suffer from imbalanced class distribution and insufficient samples within certain classes \cite{yue2022obi125,li2023towards,gao2024linking}.
Based on our analysis, this issue can be mainly attributed to the following two factors.
\begin{itemize}[topsep=0pt,itemsep=0pt]
    \item \textbf{Limited Discoveries.} Over time, the number of preserved oracle bones has become extremely limited. The corresponding discovery relies heavily on archaeological excavation, which is constrained by geographical factors and excavation techniques. Consequently, the overall number of existing oracle bones and inscriptions is relatively small, making it difficult to comprehensively cover the ancient writing system. Moreover, producing handprinted oracle characters without specialized knowledge is challenging, further complicating efforts to obtain additional samples.   
    \item \vspace{-0.15cm}\textbf{Usage Patterns.} Oracle bone scripts were primarily used in divination, which involved a relatively limited range of content and vocabulary. As a result, some characters frequently appear in divination records, while others are exceedingly rare, leading to significant imbalances in sample sizes. In addition, frequently used characters are also easier to decipher than infrequent ones, further exacerbating the imbalance in data distribution. 
\end{itemize}

In summary, oracle data scarcity and imbalance stem from historical and intrinsic script characteristics. Fig.~\ref{fig:data_distribution} presents data distributions of two datasets, i.e., OBC306 \cite{OBC306} and Oracle-AYNU \cite{icdar2019}. Specifically, OBC306 has 306 classes with 309,551 images, where the largest class has 25,898 samples, but the smallest has only one. Although majority classes (over 1,000 samples) make up 83.82\% of total samples, they represent only 25.16\% of classes \cite{OBC306}. Oracle-AYNU displays a similar, smaller pattern. Naively training on such imbalanced data often results in biased models, favoring majority classes and underperforming on minority classes. The scarcity of training samples for minority classes further complicates effective model learning. 

\subsection{Low Image Quality}\label{challenges_noise}
Characters inscribed on oracle bones often suffer significant damage, leading to poor image quality due to the following three factors \cite{ruan2024research,hust-obs,feng2015recognition}.

\begin{itemize}[topsep=0pt,itemsep=0pt]
    \item \textbf{Damage and Wear.} Oracle bones have undergone natural damage and wear, resulting in blurred or partially erased inscriptions, which directly reduces image quality. Additionally, cracks formed during divination expanded over time, causing strokes to break, deform, or disappear, further degrading the inscriptions and subsequent image quality.
    \item \vspace{-0.15cm}\textbf{Errors in Rubbing Techniques.} Many studies on oracle bones rely on rubbings, where inscriptions are transferred from the bone to paper. Uneven surfaces, cracks, and poor rubbing techniques can distort or blur characters, reducing the quality of the resulting images.
   \item \vspace{-0.15cm}\textbf{Poor Preservation Conditions.} Some oracle bones were not properly preserved after excavation and continued to deteriorate due to exposure to air and humidity. This further degraded the inscriptions, reducing the clarity of images captured later.
\end{itemize}

These factors collectively contribute to the poor quality of oracle character images. For example, in Fig.~\ref{fig:obc}, some characters are obscured by dense white areas, particularly in the upper-left and middle images. In the middle-left and middle-right images, cracks pass through the characters, which may be mistakenly identified as extra strokes, leading to recognition errors.

In summary, we find that the current OrCR faces three main challenges: variability in writing forms, limited data quantity, and poor image quality. For each challenge, we have analyzed its several possible reasons, which hopefully can help researchers understand OrCR deeply and solve these challenges to boost oracle character recognition.


\section{Datasets \& Resources}\label{Sec_dataset_resources}
\subsection{Evaluation Metrics}\label{evaluationmetrics}
To evaluate methods on OrCR, total accuracy (also known as Top-1 accuracy) is commonly used, which is formulated as $Total_{acc}=\frac{1}{N}\sum_{i=1}^{C}r_{i}$.
Here, $N$ represents the number of test samples, $C$ represents the number of classes, and $r_i$ represents the number of correctly recognized test samples of class $i$. However, many training and test sets in oracle datasets are imbalanced. Merely utilizing the total accuracy is unfair for evaluation as majority classes dominate. To address this, average accuracy is adopted, treating every class equally. It is defined as $Average_{acc}=\frac{1}{C}\sum_{i=1}^{C}\frac{r_{i}}{n_{i}}$,
where $n_i$ represents the number of test samples of class $i$. In short, for balanced oracle datasets, using total accuracy is recommended. However, for imbalanced datasets, it is preferable to leverage both total and average accuracy together for a more comprehensive evaluation.

\subsection{Datasets}\label{dataset}
To advance OrCR research, dozens of oracle datasets have been released in recent years. In addition to scanned characters, handprinted characters have also been collected to avoid the issue of image quality, and some examples of two domains are shown in Fig.~\ref{fig:oracle_charcter_examples}. We try our best to summarize all available datasets in Tab.~\ref{table:overall_datasets}, and describe their details in the following. 

\begin{enumerate}[topsep=0pt,itemsep=0pt]
    \item \textbf{Oracle-20K \cite{oracle20k}.} Oracle-20K contains 20,039 handprinted oracle character images from 261 categories.
    The dataset displays an imbalanced data distribution, with class sizes ranging from 291 images at the most to as few as 25. Additionally, it exhibits large writing variability to form significant intra-class variance.
   
    \item \vspace{-0.15cm}\textbf{SOC5519 \cite{liu2018oracle}.} SOC5519 contains 44,868 handprinted samples across 5,491 classes, nearly encompassing all known classes. Experts carefully created these images by manually tracing each line or curve from scanned rubbings. Sample numbers per class vary greatly, from a maximum of 277 to fewer than ten, with substantial intra-class variation.
    
    \item  \vspace{-0.15cm}\textbf{Oracle-AYNU \cite{icdar2019}.} Oracle-AYNU has 2,583 categories with 39,062 handprinted oracle character instances. Among these, 662 categories merely contain two instances, while the largest one contains 287 instances. The intra-class variance issue also exists in this dataset.
    
    \item \vspace{-0.15cm}\textbf{Oracle-50K/Oracle-FS \cite{han2020accv}.} Oracle-50K includes 59,081 handprinted images across 2,668 classes, with an imbalanced distribution. Derived from it, Oracle-FS supports few-shot learning with 200 classes, each having 1, 3, or 5 training samples and 20 test samples. Similarly, intra-class variation poses a challenge.\\
    Website: github.com/whhamber/Oracle-50K
    
    \item \vspace{-0.15cm} \textbf{HWOBC \cite{li2020hwobc}.} HWOBC contains 83,245 handprinted samples from historical texts, organized into 3,881 categories with balanced distribution, yet it faces challenges with intra-class variance.\\
    Website: jgw.aynu.edu.cn/home/down/index.html
    
    \item\vspace{-0.15cm} \textbf{Oracle-250/Radical-148 \cite{luxuzheng2020}.} Oracle-250 includes the 250 most frequent classes, augmented to create a balanced set of 92,160 samples. Similarly, Radical-148 selects 148 common radicals
    , with a diverse collection of data gathered from web searches and volunteers.

    \item \vspace{-0.15cm}\textbf{Ancient-3/Ancient-5 \cite{zhang2021ancient35}.} Ancient-3 and Ancient-5 each contain 1,186 classes with 39,009 handprinted oracle character samples. Ancient-3 includes early writing systems like Bronze epigraph and Chu State characters, while Ancient-5 adds Qin State and Small Seal Style characters. Both datasets face challenges with imbalanced data and intra-class variation.\\
    Website: https://figshare.com/s/c5eedcb5069c10a08830?file=44894080
    
    \item\vspace{-0.15cm} \textbf{ORCD/OCCD \cite{lin2022radical}.} ORCD is an oracle radical dataset with both rubbing source data and handprinted data, comprising 6,700 samples across 64 radical classes. OCCD includes 54,876 synthesized oracle characters by randomly combining ORCD radicals, along with 7,310 existing handprinted characters.
    
    \item \vspace{-0.15cm} \textbf{HUST-OBS \cite{hust-obs}.} HUST-OBS comprises 140,053 images, including 77,064 handprinted images of 1,588 deciphered characters and 62,989 images of 9,411 undeciphered characters, gathered from websites, books, and databases.\\
    Website: github.com/Pengjie-W/HUST-OBC
    
    \item \vspace{-0.15cm} \textbf{EVOBC \cite{evobc}.} EVOBC collects ancient characters from six historical stages: oracle, bronze, seal, Spring and Autumn, Warring States, and clerical scripts. Among them, EVOBC contains 75,681 oracle character images across 3,077 classes, sourced from various books and websites.\\
    Website: github.com/RomanticGodVAN/character-Evolution-Dataset

    \item \vspace{-0.15cm}\textbf{OBI-184 \cite{meng2018recognition}.} OBI-184 is an imbalanced dataset comprising 184 character classes and 2,000 samples. These samples are obtained by scanning and cutting rubbing inscription images from a book.
    
    \item\vspace{-0.15cm}  \textbf{OBC306 \cite{OBC306}.} OBC306 includes 309,551 instances across 306 categories, mostly scanned character images with some handprinted samples. Thus, we just categorize OBC306 as a scanned dataset. It faces severe noise, class imbalance (with the largest class at 25,898 instances and 29 classes with only one sample each), and substantial intra-class variation.\\
    Website: jgw.aynu.edu.cn/home/down/index.html
    
    \item  \vspace{-0.15cm}\textbf{OBI125 \cite{yue2022obi125}.} OBI125 is created by manually segmenting oracle characters from 1,056 rubbing images found in a book
    , resulting in an imbalanced dataset of 4,257 characters across 125 classes.\\
    Website: www.ihpc.se.ritsumei.ac.jp/obidataset.html

    \item \vspace{-0.15cm} \textbf{RCRN \cite{shi2022rcrn}.} 
    RCRN comprises a training set of 1,467 noisy-clean character image pairs and a testing set of 139 pairs, specifically designed for real-world character image denoising.\\
    Website: github.com/daqians/Noisy-character-image-benchmark

    \item \vspace{-0.15cm} \textbf{ACCID \cite{diao2023accid}.} ACCID provides character-level and radical-level annotations, covering radical categories, locations, and structural relations. It consists of 2,892 character classes with a total of 15,085 images, with 1 to 33 samples per class. Based on expert annotations, 595 radical categories have been identified, totaling 28,143 samples, with 30 to 271 samples per radical. Furthermore, each of the 14 predefined structures contains 236 to 5107 samples.

    \item  \vspace{-0.15cm}\textbf{OracleRC \cite{diao2023rzcr}.} 
    OracleRC, normalized via a denoising model \cite{shi2022charformer}, comprises 2,005 character classes. These classes can be decomposed into 202 radicals and 14 structural relations. The number of character samples per class ranges from 1 to 32.
    
    \item \vspace{-0.15cm} \textbf{Oracle-MNIST \cite{wang2024dataset}.} Oracle-MNIST contains 30,222 scanned character images across 10 commonly used classes, cropped from raw images sourced from the YinQiWenYuan website. The dataset contains serious noise and exhibits a high degree of intra-class variation.\\
    Website: github.com/wm-bupt/oracle-mnist
    
    \item  \vspace{-0.15cm}\textbf{Oracle-241 \cite{wang2022unsupervised}.} Oracle-241, collected by Anyang Normal University, consists of 78,565 instances across 241 classes, with each class including both handprinted and scanned data.\\
    Website: github.com/wm-bupt/STSN

    \item \vspace{-0.15cm} \textbf{OBIMD \cite{OBIMD}.} OBIMD includes annotation information for 10,077 pieces of oracle bones. The dataset provides detection boxes, character categories, and reading sequences for each character, making it useful for various tasks about the oracle bone script.\\
    Website: www.jgwlbq.org.cn/dt/oracleFragment
\end{enumerate}

\begin{table}[ht]
\caption{Summary of current oracle character recognition datasets. HC and SC represent handprinted and scanned characters, respectively. HR and SR represent handprinted and scanned radicals, respectively. $\text{Average}_{acc}$ and $\text{Total}_{acc}$ report the highest accuracy so far.}
\centering
\resizebox{\textwidth}{!}{
\begin{threeparttable}
\begin{tabular}{cccccccc}
\toprule
Type                       & Dataset   &Venue  & \#Classes & \#Samples &$\text{Average}_{acc}$ &$\text{Total}_{acc}$ &Public\\ \hline
\multirow{11}{*}{\textbf{HC}} 
& Oracle-20K \cite{oracle20k} &TIP 2016 & 261       & 20,039 &95.22 \cite{li2023towards} &96.17 \cite{li2023towards} &NO \\
& SOC5519 \cite{liu2018oracle} &JC 2018 & 5,491      & 44,868 &91.56 \cite{liu2018oracle} &97.16 \cite{liu2018oracle} & NO  \\
& Oracle-AYNU \cite{icdar2019} &ICDAR 2019 & 2,583     & 39,062 &86.03 \cite{li2023towards} &89.46 \cite{li2023towards} & NO\\
& Oracle-50K \cite{han2020accv}  &ACCV 2020 & 2,668     & 59,081  &-&95.68 \cite{han2020accv} & YES \\
& Oracle-FS \cite{han2020accv} &ACCV 2020  & 200       & 5,000  &-&97.59 \cite{zhao2022accv} & YES\\
& HWOBC \cite{li2020hwobc}  &JPCS 2020     & 3,881     & 83,245 &97.67 \cite{li2020hwobc} &98.17 \cite{guo2022improved} & YES \\
& Oracle-250 \cite{luxuzheng2020} &CAAI-TIS 2020 & 250       & 92,160 &- &- &NO\\
& Ancient-3/5 \cite{zhang2021ancient35} &ICCBR 2021 & 1,186     & 39,009 &- &- &YES\\
& OCCD \cite{lin2022radical} &IJDAR 2022 & 1,320     & 62,186 &- &89.20 \cite{lin2022radical} &NO \\
&HUST-OBS \cite{hust-obs}  &SD 2024    & 1,588       & 77,064 &- &- &YES\\
&EVOBC \cite{evobc}  &arXiv 2024    & 3,077       & 75,681 &- &-  &YES\\
\hline
\multirow{7}{*}{\textbf{SC}}   
& OBI-184 \cite{meng2018recognition}   &MA 2019   & 184       & 2,000 &- &92.30 \cite{meng2018recognition} &NO\\
& OBC306 \cite{OBC306}   &ICDAR 2019   & 306       & 309,551 &88.07 \cite{li2023diff} &94.12 \cite{li2023diff} &YES\\
& OBI125 \cite{yue2022obi125}  & JCCH 2022   & 125       & 4,257  &- &91.10 \cite{yue2022obi125} &YES \\
& RCRN \cite{shi2022rcrn}  & ACMMM 2022   & 362      & 1,606 &- &- &YES \\
& ACCID \cite{diao2023accid}  & ACMMM 2023   & 2,892       & 15,085  &58.17 \cite{diao2023accid} &60.28 \cite{diao2023accid} &NO\\
& OracleRC \cite{diao2023rzcr}  & IJCAI 2023   & 2,005       & - &- &61.36 \cite{diao2023rzcr} &NO\\
& Oracle-MNIST \cite{wang2024dataset}  & SD 2024   & 10       & 30,222 &93.80 \cite{wang2024dataset} &- &YES \\ 
\hline
\multirow{2}{*}{\textbf{HC+SC}}
& Oracle-241 \cite{wang2022unsupervised}  &TIP 2022 &241       & 78,565 &90.47 \cite{li2023diff}&91.11 \cite{li2023diff} &YES\\
& OBIMD \cite{OBIMD}  &arXiv 2024 &-       & 93,652 &- &- &YES\\\hline
\multirow{1}{*}{\textbf{HR}}
& Radical-148 \cite{luxuzheng2020} &CAAI-TIS 2020 & 148       & 108,989 &- &-  &NO\\
\hline
\multirow{2}{*}{\textbf{SR}}
& ACCID \cite{diao2023accid} &ACMMM 2023 & 595       & 28,143 &- &- &NO \\
&OracleRC \cite{diao2023rzcr} &IJCAI 2023& 202       &- &- &- &NO\\
\hline
\multirow{1}{*}{\textbf{HR+SR}}
& ORCD \cite{lin2022radical} &IJDAR 2022 & 64     & 6,700 &- &86.30 \cite{lin2022radical} &NO \\
\bottomrule
\end{tabular}
\label{table:overall_datasets}
\end{threeparttable}}
\end{table}

In summary, approximately 20 datasets have been developed for OrCR, greatly advancing research. As shown in Tab.~\ref{table:overall_datasets}, overall recognition accuracy for most datasets exceeds 90\%, but the lower average accuracy reveals poor performance on infrequent classes. Compared to the handprinted character (HC) datasets, recognition accuracy on scanned character (SC) datasets is suboptimal, highlighting the challenge posed by lower image quality. Recently, researchers have begun focusing on the radical components of oracle characters, constructing several datasets with radical annotations (e.g., ACCID \cite{diao2023accid}, OracleRC \cite{diao2023rzcr}, and ORCD \cite{lin2022radical}). Unfortunately, these radical datasets are relatively small, and none of them are publicly available. 

Despite numerous datasets, no standard benchmark exists for OrCR evaluation. To our knowledge, OBC306 \cite{OBC306} is widely used due to its large sample size \cite{li2021mix,huang2022acm,luo2023aggregation}, but it has limitations, such as insufficient character classes and a lack of modern Chinese labels. For research that aims to avoid image quality issues, HUST-OBS \cite{hust-obs}, which includes only handprinted characters, is recommended. Furthermore, HWOBC \cite{li2020hwobc} is a suitable choice for those seeking to avoid class imbalance.
Notably, different datasets do not share a common set of character classes. Some datasets, like OBC306 \cite{OBC306} and Oracle-AYNU \cite{icdar2019}, use only an index (i.e., ID number) for classes, without mapping to modern Chinese. As a result, oracle datasets cannot be simply merged to create a larger, unified dataset.

\subsection{Online Resources}
In addition to the formal datasets introduced in Sec.~\ref{dataset}, many online resources are available for the general public to learn about oracle bone scripts. Below, we provide a brief overview of some key resources.
\begin{itemize}[topsep=0pt,itemsep=0pt]
    \item \textbf{Yin Qi Wen Yuan.} The digital research platform specializes in the oracle bone script, offering resources such as cataloged oracle bones and academic articles. It also includes the Jingyuan Oracle Digital Platform, which regularly updates its literature database, providing a comprehensive and up-to-date academic exchange space.\\
    Website: jgw.aynu.edu.cn
    
    \item  \textbf{Guo Xue Da Shi.} An online platform dedicates to traditional Chinese culture and classical studies. It offers extensive resources on Chinese classics, including comprehensive databases on Chinese characters and texts. Notably, the platform contains the ``Oracle Bone Collection", which includes 41,956 rubbings, photographs, and facsimiles of oracle bones from Yinxu, organized into 13 volumes.\\
    Website: www.guoxuedashi.net
    
    \item  \textbf{Oracle Bone Script AI Collaborative Platform.} An online resource for the oracle bone script research and digitization. It provides a global multi-modal dataset for AI research and offers tools such as micro-trace enhancement, high-fidelity display, and glyph matching.
    \\Website: www.jgwlbq.org.cn
    
    \item \textbf{Yin Xu Oracle Bone Script Database.} This database includes 59,591 oracle bones and 143,856 divination inscriptions, with detailed transcriptions, interpretations, and annotations. It also contains an online dictionary with 3,900 character entries and 6,700 word entries, offering phonetic readings, glyph forms, and structural explanations.
    \\Website: obid.ancientbooks.cn

    \item  \textbf{Chinese Etymology.} An academic platform focuses on the origins and evolution of Chinese characters, providing information on over 100,000 ancient characters. It includes 15,000 character etymologies, 31,000 oracle bone inscriptions, 24,000 bronze inscriptions, and 49,000 seal script characters, along with Unicode standard, pronunciation, original meaning, and variant rules.
    \\Website: hanziyuan.net
\end{itemize}

\section{Methodology}\label{methodology}
In this section, we review the general progress of methodologies for oracle character recognition (OrCR). We begin by introducing general methods for recognizing oracle characters in Sec.~\ref{general}, including both traditional pattern recognition and deep learning approaches. Next, we describe specific methods aimed at addressing the challenges of writing variability, data scarcity, and low image quality in Sec.~\ref{writing_var}, Sec.~\ref{data_quantity}, and Sec.~\ref{image_quality}, respectively. Finally, we give an overall summary of methodology in Sec.~\ref{Sec_Method_summary}.

\subsection{General Oracle Character Recognition}\label{general}
Although deep learning-based methods have dominated recent advancements in OrCR, many effective works have also been developed using traditional pattern recognition techniques. To provide a comprehensive survey, we introduce both traditional pattern recognition and deep learning-based methods, and their hybrid approaches in the following.  


\textbf{1) Traditional Pattern Recognition:}
In traditional pattern recognition-based methods, manually designed features play an essential role and can be  divided into structure feature engineering and statistic feature engineering.

\textit{Structure feature engineering} methods are typically based on graph theory or topology, which parse oracle characters into specific graphical representations for recognition. For instance, Two-Level Classification (TLC) \cite{zhou1995jia} regards each oracle character as a non-directional graph and extracts its topological properties for initial recognition. Since some characters share the same topological properties, TLC further distinguishes them by extracting attributes of extensive strokes. Similarly, Graph Isomorphism Recognition (GIR) \cite{li2011isomorphism} transforms characters into undirected labeled graphs and encodes them with a quasi-adjacency matrix to compare their structures.

\textit{Statistic feature engineering} methods rely on statistical approaches to extract features from oracle character images, followed by recognition using statistical classifiers. For instance, Line Feature Recognition (LFR) \cite{Meng2017Line} extracts skeleton-based features. First, LFR utilizes Gaussian filtering and labeling to reduce noise, followed by affine transformation and thinning to extract the skeleton. Next, LFR employs a clustering method with the Hough transformation to extract line feature points. Finally, characters are recognized by measuring the distance between the line feature points of test images and templates.

\textit{Discussion:} In the structure feature engineering methods, it is essential to construct graphs to represent oracle characters. On the other hand, the statistic feature engineering methods focus on extracting distinguishable features and utilizing effective statistical classifiers. Compared to structure-based methods, statistical approaches offer greater flexibility.

\textbf{2) Deep Learning Methods:}
Recent advances in deep learning have significantly improved OrCR. In the following, we introduce these advancements, focusing on CNN-based and Transformer-based methods, respectively.

\textit{CNN-based methods:} 
Researchers have adopted various CNN architectures in OrCR \cite{alexnet,ChauhanSS24a}, where pioneer works \cite{liu2018oracle,meng2018recognition} focus on modifying basic CNNs in aspects such as activation functions, layer types, and kernel sizes. 
Subsequent studies have designed specific CNNs for OrCR \cite{gao2022enhanced,guo2022improved,luo2023aggregation}. For instance, Enhanced VGG (EVGG) \cite{gao2022enhanced} combines features of the VGG-16 \cite{vgg16} model with Inception structures, and Enhanced Inception-V3 (EIV) \cite{guo2022improved} replaces basic convolutional layers with the proposed convolutional block attention module to improve performance. However, these methods often struggle to recognize similar characters due to their limited ability to capture multi-granularity information. Multiple Attentional Aggregation Network (MAAN) \cite{luo2023aggregation} addresses this by introducing a hybrid attentional mapping unit for multi-scale feature expression and a spatial attentional aggregation unit for feature aggregation, where MAAN combines both fine-grained information from large-scale features and global semantic information from small-scale features.

Moreover, Self-Supervised Representation (SSR) \cite{du2021deep} introduces self-supervised learning to OrCR for the first time. SSR leverages pretext tasks including rotation and deformation predictions for model pre-training, followed by standard training on oracle data. SSR presents that self-supervised learning holds promise for improving OrCR performance.

Recently, researchers have started to incorporate radical information in OrCR by employing CNNs to recognize oracle radicals. Similar to modern Chinese character, radical-based methods treat an oracle character as a composition of radicals rather than a distinct character class, thus reducing vocabulary size and redundant information. 
For example, OracleNet \cite{luxuzheng2020} proposes an end-to-end model to automatically recognize radicals in oracle character images, based on CNNs and Capsule networks \cite{capsule}. Capsule units enable OracleNet to train on isolated radical data and then predict all radicals in each character image. To address the high computational cost of Capsule units, OracleNet employs transfer learning to train the CNN modules. 
Following this, Oracle Radical Extract and Recognition Framework (ORERF) \cite{lin2022radical} incorporates a detection process prior to radical recognition. In details, ORERF employs a detection network to extract radical features for radical positioning, then the detected radicals are ultimately recognized by an auxiliary classifier. 


\textit{Transformer-based methods:} With the success of Transformer models in general computer vision tasks, researchers have also adopted Transformer architectures for OrCR. Compared to CNN blocks, Transformer blocks utilize attention mechanisms to capture long-range dependencies across input data. Based on the Vision Transformers (ViT) framework \cite{vit}, Zheng et al. \cite{zheng2024ancient} propose an Improved Swin-Transformer (IST), which introduces Context-Transformer attention blocks to enhance local feature representations. 
In addition, Gan et al. \cite{gan2023graphs} introduce a Pyramid Graph Transformer (PyGT) to capture both geometric structure and spatial-temporal features using a skeleton graph.


\textit{Discussion:} Deep learning methods have significantly advanced the field of OrCR. Currently, CNN-based approaches occupy a mainstream position, where 
MAAN \cite{luo2023aggregation} achieves the highest accuracy on OBC306 \cite{OBC306}, and EIV \cite{guo2022improved} excels on HWOBC \cite{li2020hwobc}. In addition, radical-based methods with CNN models show potential for a deeper understanding of oracle characters, which could be beneficial for few/zero-shot oracle character recognition (see Sec. \ref{Sec_few_zero_learning}). However, it is challenging to obtain oracle radical information. On the other hand, Transformer-based methods have also demonstrated promise by effectively modeling visual patterns and capturing complex character structures. For example, PyGT \cite{gan2023graphs} achieves comparable performance to EIV on HWOBC \cite{li2020hwobc}. Actually, it is often unnecessary to distinguish between CNN-based or Transformer-based methods, as they are frequently integrated. For example, CNNs can be used for patch embedding in the ViT framework.

\textbf{3) Hybrid Learning Methods:}
Hybrid learning methods combine traditional pattern recognition and deep learning in OrCR. Early on, researchers integrate basic CNN models into the traditional pattern recognition frameworks, aiming to leverage the powerful representation learning ability of deep learning models. 
For example, Hierarchical Representation (HR) \cite{oracle20k} concatenates both learned features from CNN models (e.g., AlexNet \cite{alexnet}, VGGNet \cite{vgg16}, and GoogleNet \cite{googlenet}) and manually designed features (e.g., Gabor-related low-level features), followed by logistic regression for classification. HR can be regarded as a pioneering work in applying deep learning to OrCR, while subsequent studies \cite{gan2023graphs,jiang2023oraclepoints} are inspired by traditional pattern recognition ideas (e.g., the structure of oracle characters) to boost deep learning models. OraclePoints \cite{jiang2023oraclepoints} represents each oracle character using point sets and employs deep learning models to learn features for each point. PyGT \cite{gan2023graphs} is directly based on the skeleton graph. Such integration of local structure information makes deep learning models learn more robust features for OrCR, thus improving recognition performance.




\textbf{4) Summary:}
This section describes three categories of methods for OrCR, including traditional pattern recognition, deep learning, and hybrid learning methods. Traditional pattern recognition methods represent an important part of research history in OrCR. Although they may appear outdated with the advent of powerful deep learning models, these methods remain useful in certain scenarios, especially when training samples are limited. Additionally, traditional methods provide valuable insights for enhancing current deep learning approaches. Deep learning models now dominate state-of-the-art performance in OrCR, with CNN-based, Transformer-based, and hybrid CNN-Transformer models. Recently, researchers have begun exploring hybrid learning methods, where traditional pattern recognition techniques complement deep learning models, e.g. by integrating structural information to improve  performance.

\subsection{Methodology for Writing Variability}\label{writing_var}
As mentioned in Sec.~\ref{challenges_intra}, oracle characters often exhibit significant intra-class variations due to factors such as differences in handwriting or engraving styles, large spans of time and geography, and changes in the engraving medium. Methods for writing variability aim to address this issue to improve OrCR performance, primarily focusing on invariant representation learning. 

\textbf{1) Invariant Representation Learning:}
Invariant representation learning aims to extract representations that capture the essential characteristics of oracle characters while being robust to differences in styles, radicals, and structures. For instance, Topological Registration Identification (TRI) \cite{gu2016topology} argues that variant characters from the same class share similar topological structures. Therefore, characters are represented by topological graphs and recognized via a topological registration-based method.

In addition to manually defining representations, Spectral Clustering DNN (SCNN) \cite{liu2021cluster} and Isomorphism Symmetry Invariances (ISI) \cite{gao2020distinguishing} are proposed to obtain robust representations via CNN models. In details, SCNN combines a CNN model with spectral clustering, where the CNN extracts semantic features while spectral clustering aggregates different variants in each class. On the other hand, ISI introduces a two-stage recognition method that leverages the isomorphism and symmetry invariance of oracle characters. In the first stage, ISI enhances features through various transformations to train VGG-16~\cite{vgg16}, then applies prior knowledge in stage two to refine oracle variant recognition.

Instead of regarding oracle characters as entire graphics, ORERF \cite{lin2022radical} analyzes their internal structures from the perspective of radicals. ORERF is able to split radicals within oracle character images, thus reducing the overall number of variants. Additionally, since some variants arise from the same radicals appearing in different positions, detecting these positions via ORERF further reduces the number of variants.

\textbf{2) Summary:} 
Methods for writing variability mitigate the issue of severe variation within the same class by effectively learning robust representations of oracle characters, often neglected in general OrCR studies. 
Although some invariant representation learning methods have been proposed, they are still in the early stages and have not received sufficient attention. Most existing approaches rely on basic graph theory or clustering techniques, while many emerging concepts, such as prototype learning, remain largely unexplored. Moreover, with scarce samples for some oracle classes, solely pursuing invariant representation is insufficient. Even with robust feature learning, a limited number of samples may inadequately represent the full scope of a class.

\subsection{Methodology for Data Scarcity}\label{data_quantity}
As mentioned in Sec.~\ref{challenges_im}, existing oracle datasets often exhibit imbalanced class distributions, with some tail classes containing only a few samples. Compared to handprinted oracle samples, it is more difficult to obtain and annotate real-world scanned images.  
To address these issues, three strategies have been proposed for data scarcity in OrCR: imbalanced learning, few/zero-shot learning, and cross-modal learning.

\textbf{1) Imbalanced Learning:} 
Training deep learning models on imbalanced datasets often results in biased outcomes, with majority classes achieving superior performance while minority classes significantly underperform \cite{longtailsurvey}. Imbalanced learning in OrCR aims to mitigate this issue using techniques such as data augmentation and network enhancement.

\textit{Data augmentation} techniques involve artificially expanding the training set by generating additional instances of oracle characters. In OrCR, these methods generally follow trends seen in computer vision tasks, including traditional transformations, mix-up learning \cite{mixup,remix}, generative adversarial networks (GANs) \cite{WangZGJBX23,XieCSL21}, and diffusion models (DMs) \cite{YangPKZYJ24,YaoZDDZTP24}. 

Traditional transformations augment samples through random erasing, cutout, random cropping, and horizontal flipping.
Recently, Random Polygon Cover (RPC) \cite{dazheng2021random} introduces a random polygon cover algorithm, which is able to generate additional images with various polygons to mimic dense white regions in scanned images, achieving better OrCR performance. Different from transformations that rely on a single original image, mix-up learning strategies aim to leverage information from multiple images. Li et al. \cite{li2021mix} propose a Mix-up Augmentation (MA) to generate samples for minority classes by leveraging both majority and minority classes. However, MA possibly generates out-of-distribution samples due to ignoring the real data distribution.


Generative models, represented by GANs and DMs, have made great success in pattern recognition tasks, and both of them are also introduced into the field of OrCR. For instance, CycleGAN-based Data Augmentation (CDA) \cite{wang2022improving} adopts CycleGAN \cite{cyclegan} to learn the mapping between processed glyph images and scanned images, thus increasing the number of scanned samples for minority classes. Subsequently, AGTGAN \cite{huang2022acm} proposes an unsupervised GAN to translate handprinted images into scanned images, thereby supplementing the dataset with scanned oracle character images. By combining GAN and Mix-up, ADA \cite{li2023towards} introduces an end-to-end generative adversarial framework that produces synthetic data through convex combinations of all available samples in the corresponding oracle classes. However, GAN-based models often suffer from a lack of controllability and training instability \cite{difffont,WenLH021}. To overcome this issue, Diff-Oracle \cite{li2023diff}, a pioneering approach based on diffusion models, generates a diverse array of controllable oracle characters. Diff-Oracle comprises two main components for generating character images with scanned style and specific content, including a style encoder that governs the generation style based on style reference images, and a content encoder that captures specific details (e.g., glyphs) from content reference images.

\textit{Network improvement} strategies aim to enhance the architecture and training procedures of OrCR models to better handle class imbalance, achieving more equitable performance across all classes. For instance, Deep Metric Classification (DMC) \cite{icdar2019} employs a deep metric learning approach to optimize OrCR models, where a triplet loss is adopted to pull positive samples while pushing negative samples apart. MA \cite{li2021mix} also incorporates triplet loss \cite{tripletloss} with softmax on augmented samples, boosting classification accuracy.  
In addition, Decoupled Learning (DL) \cite{li2023decouple} introduces a two-stage learning method to train an unbiased DNN model for OrCR. In the first stage, the DNN model is optimized with instance-balanced sampling to establish a robust backbone, despite a biased classifier. In the second stage, a learnable weight scaling module and a KL-divergence loss are introduced to refine the classifier with class-balanced sampling. The combination of these two stages enables unbiased DNN training for improved OrCR performance.

\textit{Discussion:} 
Imbalanced learning strategies in OrCR focus on improving the accuracy of minority classes through data augmentation and network improvement techniques. Network improvement methods, such as DMC~\cite{icdar2019} and DL~\cite{li2023decouple}, utilize various loss functions and training strategies for handling imbalanced data. While effective, these methods may reduce the performance of majority classes. In contrast, data augmentation introduces additional information to enhance minority classes without compromising majority class performance. Techniques, such as mix-up strategies (e.g., MA \cite{li2021mix}) and generative models like GANs (e.g.,  AGTGAN \cite{huang2022acm} and ADA \cite{li2023towards}) and DMs (e.g., Diff-Oracle \cite{li2023diff}), have been effective in increasing the diversity and quantity of training samples, particularly for underrepresented classes. Mix-up strategies are relatively lightweight and easily integrated into existing models, whereas generative models tend to produce higher-quality samples. 

Given the importance of all classes in imbalanced learning, data augmentation remains a promising area for further exploration, particularly through generative models. Network improvement methods can complement data augmentation techniques, and combining them may yield better performance in imbalanced applications. However, the optimal integration of these two approaches remains an open research question.

\textbf{2) Few/Zero-Shot Learning:}\label{Sec_few_zero_learning}
Few/zero-shot learning in OrCR tackles the challenge of recognizing oracle characters with limited or no training instances, reflecting real-world scenarios where labeled oracle samples are often scarce. This approach encompasses two subtasks, including few-shot learning and zero-shot learning. Each subtask addresses different levels of data scarcity, aiming to improve recognition performance with minimal data.

\textit{Few-shot learning} aims to train models to recognize oracle character classes with only a few training samples, such as 1, 5, or 10 samples. Similar to imbalanced learning, few-shot learning approaches can be categorized into few-shot data augmentation and few-shot network improvement.

Some methods tackle this issue through data augmentation.  For instance, Orc-Bert \cite{han2020accv} leverages self-supervised learning on large-scale unlabelled modern handwritten Chinese characters and a few labeled oracle characters. Orc-Bert is first pre-trained on Chinese character datasets in a self-supervised manner to learn how to transform character images into stroke vectors. Few-shot oracle classes are then augmented via Orc-Bert by transforming stroke vectors. However, Orc-Bert requires a large amount of unlabelled data, and focuses merely on stroke diversity, neglecting semantic information. To overcome these limitations, Free-Form Deformation (FFD) \cite{zhao2022accv} leverages B-spline free-form deformation \cite{bspline} to randomly distort the stokes of a character without losing the overall structure, thus preserving semantic information. Moreover, FFD does not require a large additional unlabeled dataset for pre-training.

In addition to dealing with few-shot classes individually, some methods aim to promote few-shot classes while addressing imbalances \cite{huang2022acm,li2023towards,li2023diff}. For example, AGTGAN \cite{huang2022acm} augments few-shot scanned classes by translating corresponding handprinted images to scanned ones. The integration of these generated images significantly improves the accuracy of few-shot classes on scanned datasets. Furthermore, ADA \cite{li2023towards} introduces TailMix, which generates suitable few-shot samples by leveraging other similar classes. ADA expands the region of few-shot classes in the latent space, thereby improving their recognition accuracy. 

For network improvement, Siamese Similarity Network (SSN) \cite{liu2022one} modifies its network with a multi-scale fusion backbone and embedded structure for enhanced feature extraction. Meanwhile, SSN applies a soft similarity contrast loss function to increase intra-class similarity and enhance inter-class differences.

\textit{Zero-shot learning} in OrCR is designed to recognize characters that have never been seen before during training. Rather than relying on labeled samples for each class, zero-shot learning leverages auxiliary information, such as radicals, or characters from another domain, to predict unseen characters.

Radical-based methods rely on leveraging radicals to analyze oracle characters and then achieve zero-shot recognition via predefined knowledge. For instance, Diao et al.~\cite{diao2023rzcr} propose a zero-shot character recognition framework via radical-based reasoning, called RZCR.
RZCR first identifies candidate radicals and their possible structural relations from character images and then performs reasoning via a Character Knowledge Graph (CKG) for recognition. CKG stores information about characters, radicals, and structures. By updating CKG without retraining the model, RZCR is able to add new character classes. Therefore, RZCR can recognize unseen character classes by exploiting CKG with radical information and reasoning-based strategy. 
Similarly, Two-step Method for Zero-shot Oracle character Recognition (TMZOR) \cite{diao2023accid} enables zero-shot OrCR by employing character decomposition and recombination. Specifically, TMZOR first identifies the radicals and structures of characters, then utilizes a character dictionary, which contains decomposition information for zero-shot classes, i.e., radicals and structures, to perform reasoning and identifying character classes.

Another approach leverages data from the handprinted domain and translates it into the scanned domain to supplement zero-shot scanned classes. This can be referred to as zero-shot data augmentation. 
Since real-world scanned data is difficult to obtain and label, experts are invited to transcribe corresponding handprinted characters. Once the handprinted data for zero-shot classes is available, generative models such as AGTGAN \cite{huang2022acm} or Diff-Oracle \cite{li2023diff} are adopted to translate the handprinted data into realistic and diverse scanned images. These generated images maintain the same glyphs as the handprinted data and are consistent with the style of scanned data. Consequently, these images can then be used to train a recognition model, enabling accurate identification of zero-shot classes in scanned datasets.

\textit{Discussion:} Few/Zero-shot learning aims to address the challenges of minimal or absent training data. Similar to imbalanced learning, few-shot learning methods rely on both data augmentation and network improvement to handle limited data. Approaches like Orc-Bert \cite{han2020accv} and FFD \cite{zhao2022accv} augment few-shot classes based on structural transformations, significantly improving recognition accuracy. Additionally, integrating few-shot learning with imbalanced learning techniques, such as ADA~\cite{li2023towards}, generates suitable samples for few-shot classes. Network improvement methods, such as SSN \cite{liu2022one}, enhance feature extraction and loss functions to optimize recognition under few-shot conditions. Currently, data augmentation plays a key role in few-shot learning, which is explored more extensively than network improvement. However, methods like Orc-Bert and FFD rely heavily on extracting radicals from character images, which may be hindered by noises in scanned oracle characters. In the data augmentation, it is recommended to consider the writing variability for few-shot classes. 

Zero-shot learning methods like RZCR \cite{diao2023rzcr} and TMZOR \cite{diao2023accid}, leverage auxiliary information, such as radicals, to recognize unseen characters. Another direction, zero-shot data augmentation approaches like AGTGAN \cite{huang2022acm} and Diff-Oracle \cite{li2023diff}, facilitate zero-shot recognition by converting handprinted data from zero-shot classes into scanned images, addressing the scarcity of labeled scanned data. However, noisy images can impede the analysis of radicals, a frequent issue in real-world oracle character images. Currently, only handprinted data is employed for generating zero-shot scanned data, but exploring additional domains or modalities could further improve generation quality.

\textbf{3) Cross-Modal Learning:} Compared to handprinted oracle characters, it is more challenging to obtain and annotate real-world scanned images in OrCR, resulting in lower performance of OrCR in scanned characters. This challenge motivates the use of handprinted data to support the scanned domain. However, directly applying models trained on handprinted data to scanned characters often yields poor results due to domain shifts. Therefore, cross-modal learning in OrCR aims to bridge this gap by adapting models trained on the handprinted domain to perform effectively on scanned samples. 

Cross-Modal Recognition (CMR) \cite{zhang2021crossmodel} leverages both scanned and handprinted data to improve recognition performance for scanned oracle characters. CMR employs deep metric learning with a cross-modal triplet loss to map handprinted and scanned data into a shared latent space. Via adversarial learning, CMR aligns characters from both domains within the same class, ultimately recognizing scanned data with nearest neighbor classification. 

Rather than utilizing labels from scanned data, STSN \cite{wang2022unsupervised} focuses on unsupervised domain adaptation (UDA) \cite{XuWXZ23}, where scanned data lacks labels. It introduces a structure-texture separation network that disentangles features into structure and texture components, aligning handprinted and scanned data in the structure space to mitigate the negative impact of noise. By swapping texture features across domains, STSN translates features between handprinted and scanned data, thereby enhancing the recognition performance on scanned data. However, optimizing GANs can be complex, and cross-entropy loss may limit feature discriminability. To address these issues, Unsupervised Discriminative Consistency Networks (UDCN) \cite{wang2024unsupervised} is proposed. UDCN utilizes pseudo-labeling to integrate semantic information into the adaptation process and enforce augmentation consistency, ensuring consistent predictions of scanned samples under various perturbations. Additionally, UDCN introduces an unsupervised transition loss to improve feature discriminability in the scanned domain by optimizing both between-class and within-class transition probabilities.

\textit{Discussion:} 
Cross-modal learning offers significant advantages by leveraging abundant handprinted data to improve recognition performance on scarce scanned data. These methods reduce the need for massive scanned data and time-consuming image labeling, making them practical for real-world applications. Furthermore, they enhance the generalization ability of models, enabling them to handle domain shifts effectively and maintain high recognition accuracy across different data distributions.

Despite these benefits, current methods do not fully exploit the intrinsic properties of oracle characters, such as radicals, which presents a valuable area for further exploration. Moreover, existing works mainly focus on close-set scenarios. Further research could extend these methods to address more complex distribution shifts, such as partial and open-set UDA \cite{wang2024unsupervised}, which are more relevant for practical applications.

\textbf{4) Summary:}
This section introduces recent progress in addressing data scarcity in OrCR, attracting significant attention with promising results. Researchers have explore three main directions, including imbalanced learning, few/zero-shot learning, and cross-modal learning.
Imbalanced learning utilizes data augmentation and network improvement to boost minority class recognition. Data augmentation is worth exploring since it often boosts performance without sacrificing majority class performance. Few/zero-shot learning improves recognition despite minimal data using methods like structural transformations and generative models. Cross-modal learning leverages handprinted data to enhance scanned data recognition without requiring extensive labeled scanned samples. Despite these advances, challenges still remain, such as handling noisy images and addressing more complex scenarios like open-set recognition. Future work could focus on better integrating character properties, such as radicals, with generative models to further improve recognition performance.


\subsection{Methodology for Low Image Quality}\label{image_quality}
Methods for low image quality in OrCR refer to developing methods to accurately recognize oracle characters from scanned images that contain various forms of noise, as mentioned in Sec.~\ref{challenges_noise}. Such noise significantly distorts or obscures important details of the characters. There are two main approaches for addressing noise issues, including image denoising and noise simulation. 

\textbf{1) Image Denoising:}
Image denoising aims to reduce noise from raw images before inputting them into a recognition model. Gu et al. \cite{gu2010restoration} propose a method for restoring characters in scanned images using a combination of Poisson distribution and fractal geometry. However, applying such methods requires prior knowledge of the noise type and level, which is challenging to obtain in real-world image restoration scenarios. Recently, researchers have employed deep learning models for this task in blind noise reduction scenarios, where the noise type and level are unknown. For instance, Charformer \cite{shi2022charformer} introduces a generic end-to-end framework based on glyph fusion and attention mechanisms to precisely recover character images without altering their glyphs. Charformer incorporates a parallel target task to capture additional information and inject it into the image-denoising backbone, thus maintaining the consistency of character glyphs during the denoising process. Attention-based networks are utilized for global-local feature interaction, aiding in blind denoising and enhancing overall performance. Another work Real-world Character Restoration Network (RCRN) \cite{shi2022rcrn} restores degraded character images by utilizing character skeleton information and scale-ensemble feature extractions. In details, RCRN includes a skeleton extractor and character image restorer. The skeleton extractor preserves the structural consistency of the character and normalizes complex noise. Then, the character image restorer reconstructs clean images from the degraded images and their skeletons.

\textit{Discussion:} In image denoising, maintaining the integrity of original glyphs is crucial. Both Charformer~\cite{shi2022charformer} and RCRN~\cite{shi2022rcrn} have demonstrated strong performance in this regard.
However, they face challenges under extreme noise conditions, where image reconstruction may fail to fully restore details, and residual artifacts may remain.

\textbf{2) Noise Simulation:}
Noise simulation methods introduce artificial noise into training data to improve model robustness against real-world noise. 
RPC \cite{dazheng2021random} addresses white region noise in scanned data by proposing a random polygon cover algorithm to simulate such damage during training. Additionally, AGTGAN \cite{huang2022acm} generates natural background noise by translating handprinted images into scanned images. Compared to AGTGAN, Diff-Oracle~\cite{li2023diff} offers more controllable and diverse noise generation. Given a scanned image as reference style, Diff-Oracle effectively captures stylistic attributes, such as dense white regions and missing edges, to generate realistic noise in synthesized images. Training with these images from generative models enhances the model robustness, allowing effective recognition of noisy character images. 

\textit{Discussion:} Noise simulation methods enhance recognition robustness by exposing models to various forms of artificial noise during training. This exposure helps models generalize better to real-world noisy conditions. For example, 
Diff-Oracle \cite{li2023diff} demonstrates the optimal accuracy on Oracle-241 \cite{huang2022acm} and OBC306 \cite{OBC306}. However, noise simulation methods might struggle to accurately replicate the full complexity of real-world noise. Generated noise may not accurately capture all variations, potentially limiting model effectiveness. In addition, over-generating noisy images can negatively impact the recognition of clean images, further complicating the process.

\textbf{3) Summary:}
Scanned oracle characters often suffer from low image quality. Two primary strategies address this issue: image denoising and noise simulation. Image denoising aims to restore degraded images to improve recognition performance, while noise simulation enhances model robustness by exposing models to diverse noises during training. Although both approaches have shown promising results, research on addressing low-quality oracle images remains under-explored. Future work could benefit from integrating both methods for a more comprehensive solution. Combining advanced denoising techniques with noise simulation may enhance model performance under extreme noise conditions, leading to more reliable oracle character recognition.




\subsection{Overall Summary}
\label{Sec_Method_summary}
In summary, this section has presented recent progress in OrCR, a field that has garnered considerable attention. Based on our analysis, OrCR faces three primary challenges. We have detailed advancements addressing each specific challenge, noting that while significant progress has been made, no challenge has been fully resolved. Furthermore, emerging trends with deep learning models suggest promising directions, as discussed in Sec.~\ref{future_works}. For a comprehensive overview, Tab.~\ref{tab:summary_methods} consolidates the key OrCR methods covered.

\begin{table}[ht]
\caption{Summary of representative oracle character recognition methods. ``General” indicates general recognition methods, including traditional pattern recognition (TPR), deep learning methods (DLM), and hybrid learning methods (HLM). ``MWV" indicates methodology for writing variability by invariant representation learning (IRL). ``MDS" represents methodology for data scarcity, including imbalanced learning (IL), few/zero-shot learning (FZSL), and cross-modal learning (CML). ``MLIQ" indicates methodology for low image quality, including image denoising (ID) and noise simulation (NS). }
\centering
\label{tab:summary_methods}
\renewcommand{\arraystretch}{1.2} 
\resizebox{0.9\textwidth}{!}{
\begin{tabular}{>{\centering\arraybackslash}p{2cm} >{\centering\arraybackslash}p{3cm} >{\centering\arraybackslash}p{8cm}}
\toprule
Category  & Sub-Category  & Representative Work \\ \midrule
\multirow{4}{*}{\textbf{General}} & TPR    & TLC~\cite{zhou1995jia}, GIR~\cite{li2011isomorphism}, LFR~\cite{Meng2017Line} \\ \cline{2-3} 
                         & \multirow{2}{*}{DLM}    & OracleNet~\cite{luxuzheng2020}, ORERF~\cite{lin2022radical}, EVGG~\cite{gao2022enhanced}, EIV~\cite{guo2022improved}, MAAN~\cite{luo2023aggregation}, PyGT~\cite{gan2023graphs} \\ \cline{2-3}
                         & HLM    & HR~\cite{oracle20k}, OraclePoints~\cite{jiang2023oraclepoints}, PyGT~\cite{gan2023graphs} \\ \midrule
\textbf{MWV}                     & IRL     & TRI~\cite{gu2016topology}, ISI~\cite{gao2020distinguishing}, SCNN~\cite{liu2021cluster} \\ \midrule
\multirow{5}{*}{\textbf{MDS}}    & \multirow{2}{*}{IL}      & DMC~\cite{icdar2019}, MA~\cite{li2021mix}, AGTGAN~\cite{huang2022acm}, ADA~\cite{li2023towards}, Diff-Oracle~\cite{li2023diff}, DL~\cite{li2023decouple} \\ \cline{2-3}
                         & \multirow{2}{*}{FZSL}    & FFD~\cite{zhao2022accv}, AGTGAN~\cite{huang2022acm}, RZCR~\cite{diao2023rzcr}, TMZOR~\cite{diao2023accid}, ADA~\cite{li2023towards}, Diff-Oracle~\cite{li2023diff} \\ \cline{2-3}
                         & CML     & CMR~\cite{zhang2021crossmodel}, STSN~\cite{wang2022unsupervised}, UDCN~\cite{wang2024unsupervised} \\ \midrule
\multirow{2}{*}{\textbf{MLIQ}}   & ID      & PDFG~\cite{gu2010restoration}, Charformer~\cite{shi2022charformer}, RCRN~\cite{shi2022rcrn} \\ \cline{2-3}
                         & NS      & RPC~\cite{dazheng2021random}, CDA~\cite{wang2022improving}, AGTGAN~\cite{huang2022acm}, Diff-Oracle~\cite{li2023diff} \\ 
\bottomrule
\end{tabular}}
\end{table}

\section{Related Tasks}\label{applications}
OrCR is essential for studying oracle bone scripts. Here, we discuss its relationship with seven related tasks: oracle character decipherment, oracle character evolution, oracle character image retrieval, oracle character visual guide, oracle character detection, oracle character segmentation, and oracle bone joining.

\subsection{Oracle Character Decipherment (OrCD)}
OrCD seeks to interpret unknown oracle characters and link them to modern Chinese characters. It requires both identifying individual unknown oracle characters and understanding their contextual meanings within the broader oracle characters. In this process, OrCR supports OrCD by recognizing known oracle characters, expediting the identification of frequently recurring or difficult-to-recognize characters. Recognized characters and their context aid in inferring unknown meanings, as contextual relationships are crucial for interpretation. However, current OrCD efforts \cite{gao2022image,wang2024puzzle,guan2024deciphering} mainly focus on image processing for translating oracle characters to modern Chinese, often overlooking valuable contextual information. Thus, developing OrCD methods that incorporate context is a promising research direction deserving further exploration.

\subsection{Oracle Character Evolution (OrCE)}
OrCE is a system designed to study the evolution of Chinese characters from the Shang dynasty to the present by analyzing structural and morphological changes from deciphered oracle characters to subsequent scripts. In this context, OrCR plays a vital role by offering deciphered characters and evaluation metrics for OrCE. While OrCD and OrCE share similarities, their focus diverges: OrCD focuses on character meanings, whereas OrCE emphasizes relationships and evolutionary patterns across dynasties. Oracle characters are ideographic, with meanings conveyed through stable components, even as forms change. Current OrCE research \cite{chang2022sundial,wang2022study} mainly examines whole-character changes, but limited one-to-one evolutionary data complicates understanding. Thus, exploring relationships and structural changes within oracle character components is a promising area for further study in script evolution.

\subsection{Oracle Character Image Retrieval (OrCIR)}
OrCIR retrieves oracle character images based on a query image, identifying similar or identical characters \cite{liu2020oracle,hu2024component,ding2024oracle}. Unlike OrCR, which labels characters, OrCIR aims to find visually similar characters from a query. Both share optimization goals like improving image quality and feature extraction, so advancements in OrCR could benefit OrCIR. However, most research in OrCIR has targeted known character retrieval, with limited focus on representing unknown characters—a topic warranting further investigation.
 

\subsection{Oracle Character Visual Guide (OrCVG)}
Oracle characters are challenging for modern audiences due to centuries of evolution. OrCVG addresses this by generating visual representations and interpretative guides based on oracle glyphs to aid public understanding. This emerging field, driven by advances in generative AI, offers an educational tool through visual aids and explanations. Recently, Qiao et al.~\cite{qiao2024making} propose the GenOV (Generative Oracle Visuals) framework, which combines a large vision-language model (e.g., QWEN-VL~\cite{Qwen-VL}) for reasoning and a text-to-image model (e.g., GLIGEN~\cite{LiLWMYGLL23}) for visual generation to create accurate visual guides for oracle characters. Unlike OrCR, which automates recognition to label oracle characters, OrCVG provides semantic understanding and illustrations, adding a new dimension to oracle character research.


\subsection{Oracle Character Detection (OrCDe)}
As outlined in Sec.~\ref{intro}, the recognition process includes detection and then recognition. However, OrCDe \cite{jiao2021module,fu2024detecting,fu2024shape} is less studied than OrCR and faces challenges like noise, image degradation, and overlapping inscriptions, resulting in suboptimal performance. Advanced attention mechanisms and multi-scale detection models \cite{internimage,yolov10} could improve robustness, while few-shot detection could address the scarcity of labeled data for rare characters. Researchers could also explore joint detection and recognition \cite{yang2020oracledetection,fujikawa2023recognition} to optimize the entire pipeline.

\subsection{Oracle Character Segmentation (OrCS)}
OrCS, which is essential for isolating characters prior to recognition, receives even less attention than detection. Ge et al. \cite{ge2021oracle} apply the BoxInst method \cite{boxinst} that is originally developed for general instance segmentation to address this task, yet results are limited by challenges like incomplete or eroded inscriptions and high dependence on manual annotations. Future research should prioritize models that handle degraded inscriptions and reduce annotation costs through weakly-supervised or unsupervised approaches \cite{moco,simclr}. Additionally, advanced image enhancement techniques could improve segmentation accuracy for incomplete characters, benefiting subsequent OrCR tasks.

\subsection{Oracle Bone Rejoining (OrBR)}
OrBR reconstructs fragmented bones,  aiding in complete inscription interpretation and supporting OrCR. Current methods mainly assess contour similarity between fragments \cite{zhang2022datadriven,zhang2022internal,yuan2023sffsiam}, but challenges remain with noise, degradation, and limited annotated data. In the future, researchers could enhance contextual understanding by leveraging spatial relationships and integrating visual and textual information, like residual characters and homographs. Additionally, treating OrBR and OrCR as a unified task may create a feedback loop, improving both processes and advancing oracle inscription understanding.

\section{Future Work}\label{future_works}
Recent advances in pattern recognition and deep learning have significantly improved oracle character recognition (OrCR) performance. However, several key areas still require future exploration. 
This section highlights important directions for future research, focusing on data, class sets, and model learning: Robust OrCR for data issues, Open-Set OrCR for class sets, and Self-Supervised OrCR for learning. Given the current focus on large models, we also explore their potential in OrCR. This section aims to inspire further advancements and provide new perspectives for the OrCR research community.

\subsection{Robust Oracle Character Recognition}
Due to the difficulty of annotating oracle characters, some oracle datasets may suffer from label noise \cite{yue2022obi125}. However, most OrCR methods assume that all labels are clean, leading to poor model robustness in practical applications. Therefore, robust oracle character recognition seeks to improve model performance despite inconsistencies in the data.

\subsection{Open-Set Oracle Character Recognition}
In real applications, oracle characters often follow an open-ended class distribution, with many characters undeciphered. Open-set oracle character recognition aims to recognize known classes while managing unknown ones, facing challenges in accurately distinguishing between them and balancing sensitivity to unknown classes with accuracy on known classes. This is further complicated by dataset imbalance. DMC \cite{icdar2019} explores this with deep metric learning. Combining OrCR with oracle character decipherment research \cite{guan2024deciphering,wang2024puzzle} may enhance open-set recognition. Additionally, insights from open-set and open-set imbalanced learning \cite{osl,osil1,osil2} could support this area.


\subsection{Self-Supervised Oracle Character Recognition}
Many oracle character images lack annotations, leaving much of the available unlabeled data underutilized in OrCR. Incorporating self-supervised learning can effectively leverage them to enhance feature extraction, thereby improving model generalization and recognition performance \cite{moco}. Although SSR \cite{du2021deep} attempts this approach with simple transformation predictions, it remains a shallow application. Advanced self-supervised methods, such as contrastive learning \cite{moco,simclr}, hold significant potential for advancing this field.

\subsection{Oracle Character Recognition with Large Models}
Recent advances in large models have transformed fields like computer vision \cite{sd,dalle}, NLP \cite{gpt3,palm}, and other domains \cite{clip,flamingo}, excelling in complex data analysis due to their extensive parameters and layers. These models capture intricate patterns, processing large datasets with high accuracy. Applying such models to OrCR could prove advantageous; building an OrCR-specific large model or fine-tuning pre-trained models could enhance performance, even with limited data. Diff-Oracle \cite{li2023diff} makes an initial try by utilizing pre-trained ControlNet \cite{controlnet} as a basis model for oracle character generation, while Diao et al. \cite{qiao2024making} leverage GPT \cite{gpt3} for visual reasoning tasks related to oracle glyphs.


\section{Conclusion}\label{conclusion}
This paper presents a comprehensive survey of oracle character recognition (OrCR), providing an overview of recent advancements in the field. We begin by introducing the key challenges and datasets to help readers understand the current landscape of OrCR research. We then review existing pattern recognition and deep learning methods for OrCR, discussing their strengths and limitations in detail. Next, we explore several related tasks of OrCR and highlight promising directions for future research in methodologies and task settings. This survey aims to deepen the research community's understanding of OrCR and inspire new advancements to drive further progress in this evolving field.

\section*{Acknowledgements}
The work was partially supported by the following: National Natural Science Foundation of China under No.92370119, No.62276258, and No.62376113, and open program of Henan Key Laboratory of Oracle Bone Inscription Information Processing (AnYang Normal University) under No.OIP2024H004.

\bibliographystyle{elsarticle-num}
\bibliography{reference_abb}

\end{document}